\pdfoutput=1

\documentclass[11pt]{article}

\usepackage[preprint]{acl}
\usepackage{amsmath}
\usepackage{xcolor}
\usepackage{booktabs}
\usepackage{array}
\usepackage{amssymb}
\usepackage{geometry}
\usepackage{multirow}
\usepackage{graphicx}
\usepackage{times}
\usepackage{tabularx}
\usepackage{latexsym}
\usepackage{xcolor}
\definecolor{softred}{rgb}{0.9, 0.4, 0.4}
\definecolor{softblue}{rgb}{0.4, 0.4, 0.9}
\usepackage[T1]{fontenc}

\usepackage[utf8]{inputenc}

\usepackage{microtype}

\usepackage{inconsolata}
\usepackage[linesnumbered,ruled,vlined]{algorithm2e}
\usepackage{graphicx}

\title{Warmup Generations: A Task-Agnostic Approach for Guiding Sequence-to-Sequence Learning with
Unsupervised Initial State Generation }

\author{
    Senyu Li$^{1,2}$\qquad 
    Zipeng Sun$^{1,2}$\qquad 
    Jiayi Wang$^3$\qquad \\
    \textbf{Xue Liu}$^{1,2}$\qquad 
    \textbf{Pontus Stenetorp}$^3$\qquad 
    \textbf{Siva Reddy}$^{1,2,4}$\qquad 
    \textbf{David Ifeoluwa Adelani}$^{1,2,4}$\\
    $^1$Mila - Quebec AI Institute, $^2$McGill University, $^3$University College London,\\  $^4$Canada CIFAR AI Chair \\
    \texttt{\{senyu.li,  siva.reddy, david.adelani\}@mila.quebec }\\
    \texttt{\{zipeng.sun, xueliu\}@mail.mcgill.ca}\\
    \texttt{\{jiaywang, p.stenetorp\}@cs.ucl.ac.uk}\\
    }

\begin{document}

\maketitle
\begin{abstract}
Traditional supervised fine-tuning (SFT) strategies for sequence-to-sequence tasks often train models to directly generate the target output.
Recent work has shown that guiding models with intermediate steps—such as keywords, outlines, or reasoning chains—can significantly improve performance, coherence, and interpretability. However, these methods often depend on predefined intermediate formats and annotated data, limiting their scalability and generalizability. In this work, we introduce a task-agnostic framework that enables models to generate intermediate ``warmup'' sequences. These warmup sequences, serving as an initial state for subsequent generation, are optimized to enhance the probability of generating the target sequence without relying on external supervision or human-designed structures. Drawing inspiration from reinforcement learning principles, our method iteratively refines these intermediate steps to maximize their contribution to the final output, similar to reward-driven optimization in reinforcement learning with human feedback. Experimental results across tasks such as translation, summarization, and multi-choice question answering for logical reasoning show that our approach outperforms traditional SFT methods, and offers a scalable and flexible solution for sequence-to-sequence tasks\footnote{We will release our code after the paper is published.}.

\end{abstract}

\section{Introduction}

Recent advancements in large-scale pre-trained language models (LLMs), such as T5 \cite{DBLP:journals/corr/abs-1910-10683}, GPT 
 \cite{brown2020languagemodelsfewshotlearners}, and LLaMA \cite{touvron2023llamaopenefficientfoundation}, have significantly improved performance on both predictive tasks (e.g., multi-choice question answering) and generative tasks (e.g., machine translation and summarization). These models have demonstrated exceptional capabilities in generating coherent and contextually relevant outputs by modelling dependencies across long sequences of data. Despite their successes, traditional 
supervised fine-tuning (SFT) methods
 for such models
 often focus on directly generating the target output without leveraging the benefits of intermediate steps or initial guidance \cite{sutskever2014sequencesequencelearningneural}.

Research has shown that guiding models with intermediate steps, such as outlines, keywords, or reasoning chains, can significantly improve performance, coherence, and interpretability across tasks \cite{wang2022rationaleaugmentedensembleslanguagemodels, creswell2022faithfulreasoningusinglarge}.
For instance, hierarchical frameworks for tasks such as story generation \cite{fan-etal-2019-strategies} and summarization \cite{amplayo2020unsupervisedopinionsummarizationcontent} often first generate high-level structures, such as outlines or reasoning steps, before producing detailed outputs. 
These approaches highlight the utility of intermediate guidance in organizing complex tasks. Similarly, chain-of-thought (COT) \cite{wei2023chainofthoughtpromptingelicitsreasoning} reasoning for predictive tasks extends this concept by demonstrating the value of logical steps by decomposing complex predictive tasks into explicit logical steps, demonstrating how structured reasoning between inputs and outputs can improve model performance.
\begin{figure}
    \centering
    \includegraphics[width=0.5\textwidth, height=0.11\textheight]{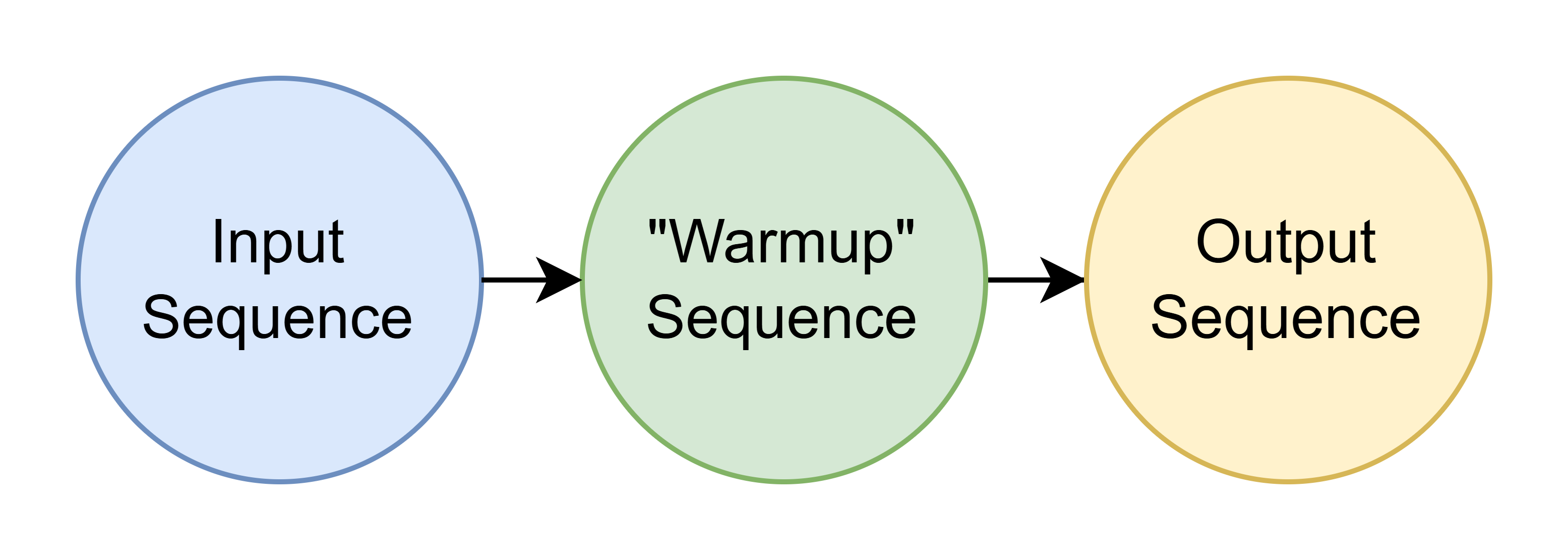}
    \vspace{-7mm}
    \caption{\textbf{High-level workflow of Warmup Generations. } The input is first used to generate an intermediate ``warmup'' sequence, which acts as a guiding context to improve the generation of the final target output for sequence-to-sequence tasks.}
    \label{fig:intuitve}
\end{figure}
However, these approaches heavily rely on predefined intermediate formats and annotated data, which are costly to create due to human annotation efforts and often task-specific, thus limiting their scalability and adaptability to broader applications.

In this work, we address these limitations by introducing a framework that enables models to generate an initial state, which we refer to as ``warmup sequence.'' These warmup sequences act as preparatory steps, priming the model for the main generation task. Drawing inspiration from reinforcement learning (RL) principles, our method treats these steps as actions within a reward-driven framework, optimizing them to maximize their utility in improving the quality and coherence of the final target output. Importantly, this process operates without relying on predefined formats or external annotations, making it adaptable to a wide range of tasks and model architectures.
This approach eliminates dependence on annotated data for intermediate steps, achieves generalization across tasks, and unifies the optimization of intermediate and final outputs, leading to improved 
final performance of the models.

Through experiments, we demonstrate that our method improves output quality across translation, summarization,
and multi-choice question answering for logical reasoning, and is compatible with various model architectures, including encoder-decoder models like T5 \cite{2020t5} and mT5 \cite{xue2021mt5massivelymultilingualpretrained}, as well as decoder-only models like Llama \cite{touvron2023llamaopenefficientfoundation}. In addition, our method is simple to implement, requiring only about 10 additional lines of codes, without modifications
to existing model architectures or reliance on task-specific annotations, and is grounded in a solid theoretical framework. These contributions establish an approach where models can autonomously discover and leverage a helpful
initial state that increases the probability of the target sequence 
across diverse tasks to enhance the quality of the final generation.

\begin{figure*}
    \centering
    \includegraphics[width=1\textwidth, height=0.12\textheight]{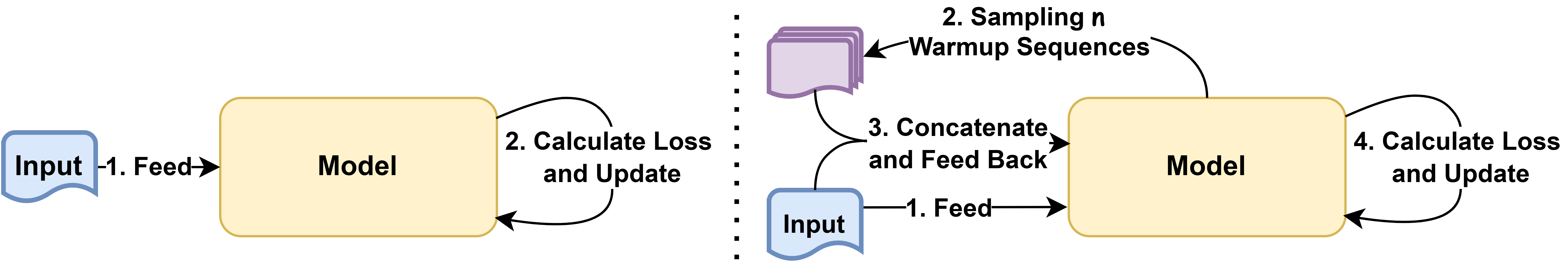}
    \caption{\textbf{Comparison of the traditional supervised fine-tuning methods (left) and our proposed method (right)}. The traditional method directly optimizes the mapping from input to output using annotated data, while our method dynamically generates and optimizes warmup sequences to guide the final output.}
    \label{fig:workflow}
\end{figure*}

\section{Related work}
Guiding generative models with intermediate steps has been widely explored to enhance coherence, interpretability, and task performance. Existing approaches can be categorized into explicit human-readable intermediate steps, and structured weakly supervised intermediate steps.

\paragraph{Human readable intermediate steps}
Plan-and-Write \cite{Yao_Peng_Weischedel_Knight_Zhao_Yan_2019} introduces storyline-based planning, where a model first generates a structured sequence of key events before expanding them into a full story, improving coherence and creativity. 
\citeauthor{amplayo2020unsupervisedopinionsummarizationcontent} \citeyearpar{amplayo2020unsupervisedopinionsummarizationcontent} employ content planning, explicitly modelling aspect and sentiment distributions to guide summary generation, thereby enhancing readability and informativeness.

Similarly, \citeauthor{wolfson-etal-2022-weakly} \citeyearpar{wolfson-etal-2022-weakly} propose Question Decomposition Meaning Representations (QDMR), which break down complex questions into sequences of reasoning steps. These decompositions serve as an explicit intermediate representation, improving interpretability and guiding Text-to-SQL parsing by systematically mapping natural language queries to SQL.
\citeauthor{baziotis-etal-2019-seq} \citeyearpar{baziotis-etal-2019-seq}  introduce a sequence-to-sequence-to-sequence model (SEQ³), where the intermediate step is a compressed version of the input sentence, explicitly represented in natural language.


\paragraph{Structured intermediate steps}
Some models introduce structured but weakly supervised intermediate steps, where the intermediate representations are partially interpretable but not explicitly labelled during training. \citeauthor{cheng-etal-2017-learning} \citeyearpar{cheng-etal-2017-learning} generate predicate-argument structures, which serve as an intermediate step in semantic parsing. Unlike explicit intermediate representations, these structures are learned through optimization-based search rather than direct supervision.
Similarly, \citeauthor{jambor-bahdanau-2022-lagr} \citeyearpar{jambor-bahdanau-2022-lagr} propose Label Aligned Graphs (LAGr), where models predict node and edge labels to construct structured meaning representations aligned with input text, improving systematic generalization in semantic parsing. These representations enhance compositional generalization but still depend on predefined structural mappings.
\citeauthor{herzig2021unlockingcompositionalgeneralizationpretrained} \citeyearpar{herzig2021unlockingcompositionalgeneralizationpretrained} introduce intermediate representations that transform meaning representations (e.g., SPARQL or SQL queries) into structured forms that improve compositional generalization while maintaining reversibility. 
While these methods balance interpretability and generalization, they still rely on task-specific constraints rather than fully flexible intermediate representations.

\paragraph{Reinforcement learning in NLP}
RL has also been applied in NLP to optimize model generation beyond traditional supervised learning for text summarization~\citep{paulus2018a}, dialogue generation \cite{li-etal-2016-deep} and machine translation \cite{wu-etal-2018-study}. More recently, Reinforcement Learning from Human Feedback (RLHF) \cite{NIPS2017_d5e2c0ad} has been instrumental in aligning large-scale language models with human preferences, demonstrating the effectiveness of RL-based fine-tuning. While RL provides a strong optimization framework, it does not inherently generate structured intermediate representations, instead refining model behaviour through reward-based learning.



In this paper, we learn intermediate steps freely, without predefined formats, constraints, search procedures, external supervision, annotated datasets or task-specific designs.
Inspired by RL principles, our method integrates intermediate step/initial state generation and final output optimization into a unified framework. 
Our approach generalizes across tasks such as translation, summarization, and multi-choice question answering logical reasoning, and architectures, providing a flexible, scalable, and theoretically grounded solution for improving the quality of generation.

\section{Formulation and Derivation}
We reformulate the process of text generation by assuming that given a specific \textbf{input} \(x\) and the \textbf{target text} \(y_{\text{target}}\), there exists an intermediate sequence, or the \textbf{initial state} \(c_{\text{init}}\) preceding \(y_{\text{target}}\), where \(\text{length}(c_{\text{init}}) \geq 0\). To be more specific, \( c_{\text{init}} = \{c_1, c_2, \ldots, c_k\} \) is a sequence of tokens, where \( k \geq 0 \). The intermediate sequence \( c_{\text{init}} \) serves as a latent variable that conditions the generation of \( y_{\text{target}} \). When \( k = 0 \), \( c_{\text{init}} \) is an empty sequence, reducing the framework to the traditional sequence-to-sequence paradigm:
\vspace{-2mm}
\[
P(y_{\text{target}}|x) = \sum_{c_{\text{init}}} P(c_{\text{init}}, y_{\text{target}}|x)
\]

\vspace{-2mm}
Using the chain rule, this can be decomposed as:
\vspace{-2mm}
\[
P(y_{\text{target}}|x) = \sum_{c_{\text{init}}} P(y_{\text{target}}|c_{\text{init}}, x)P(c_{\text{init}}|x)
\]
\vspace{-2mm}
Which can be rewritten in the form:
\vspace{-3mm}

\[
P(y_{\text{target}}|x) = \mathbb{E}_{c_{\text{init}} \sim P(c_{\text{init}}|x)}[P(y_{\text{target}}|c_{\text{init}}, x)]
\]

Our objective is to maximize the probability of the target sequence \(y_{\text{target}}\) given the input \(x\). Traditionally, the loss for maximizing \(P(y_{\text{target}}|x)\) is:

\vspace{-5mm}
\[
L_{y_{\text{target}}} = -\log(P(y_{\text{target}}|x))
\]

Which is equivalent to:

\[
L_{y_{\text{target}}} =  -\log(\mathbb{E}_{c_{\text{init}} \sim P(c_{\text{init}}|x)}[P(y_{\text{target}}|c_{\text{init}}, x)])
\]
where \(c_{\text{init}}\) represents any possible initial state conditioning \(y_{\text{target}}\). This expectation implies maximizing \(P(y_{\text{target}})\) across all initial states, weighted by their probability \(P(c_{\text{init}}|x)\).

\section*{Reward-Based Initial State Optimization}

Since we lack labels for \(c_{\text{init}}\), we train the model to generate \(c_{\text{init}}\) using reward-based optimization. A good initial state \(c_{\text{init}}\) increases the probability of \(y_{\text{target}}\), while a poor \(c_{\text{init}}\) reduces it. The reward \( R(c_{\text{init}}) \) quantifies the quality of \( c_{\text{init}} \) in terms of its contribution to generating \( y_{\text{target}} \) given \( x \):
\vspace{-2mm}
\[
R(c_{\text{init}}) = P(y_{\text{target}}|c_{\text{init}}, x)
\]

\vspace{-2mm}
Under a RL framework, we aim to maximize:

\vspace{-2mm}
\[
\mathbb{E}_{c_{\text{init}} \sim P(c_{\text{init}}|x)}[R(c_{\text{init}})]
\]
which is equivalent to:

\vspace{-2mm}
\[
\mathbb{E}_{c_{\text{init}} \sim P(c_{\text{init}}|x)}[P(y_{\text{target}}|c_{\text{init}}, x)]
\]

The loss for training \(c_{\text{init}}\) generation is defined as the negative log of the expected reward:

\vspace{-3mm}
\begin{align*}
L_{c_{\text{init}}} &= -\log\big(\mathbb{E}_{c_{\text{init}} \sim P(c_{\text{init}}|x)}[R(c_{\text{init}})]\big) \\
&= -\log\big(\mathbb{E}_{c_{\text{init}} \sim P(c_{\text{init}}|x)}[P(y_{\text{target}}|c_{\text{init}}, x)]\big)
\end{align*}


\begin{algorithm}

\caption{Warmup Generations}
\label{algo:1}
\KwIn{Training Data,  Maximum Epochs $E$, Number of Samples $n$, Model Parameters $\theta$}
\KwOut{Trained $\theta$}

\SetKwBlock{Initialize}{Initialize:}{}
\Initialize{
    Model $\theta$ with pretrained weights\;
}

\For{epoch $t = 1$ to $E$}{
    \For{each input $x$ in Training Data}{
        
        Initialize total loss $\mathcal{L}_{\text{total}} \leftarrow 0$\;
        \For{$i = 1$ to $n$}{
           
            Sample $c_{\text{init}}^{(i)} \sim P(c_{\text{init}}|x; \theta)$ \;

            Compute loss $\mathcal{L}^{(i)}$ for $y_{\text{target}}^{(i)}$\ conditioned on $c_{\text{init}}^{(i)}$ and $x$\;
            $\mathcal{L}_{\text{total}} \leftarrow \mathcal{L}_{\text{total}} + \mathcal{L}^{(i)}$\;
        }
       
        $\mathcal{L}_{\text{avg}} \leftarrow \mathcal{L}_{\text{total}} / n$\;

        Update $\theta$ using gradient descent\;
    }

}
\Return{Trained $\theta$}

\end{algorithm}

As we can observe, this formulation aligns the optimization of \(y_{\text{target}}\) and \(c_{\text{init}}\) under the same loss function.
\vspace{-2mm}
\[
L_{c_{\text{init}}} = L_{y_{\text{target}}}
\]

Directly minimizing $ L_{c_{\text{init}}}$ or $L_{y_{\text{target}}}$ is computationally infeasible due to numerical underflow for long sequences. Instead, we minimize the expected cross-entropy loss based on Jensen's inequality:




\vspace{-5mm}

\begin{align*}
-\log\big(&\mathbb{E}_{c_{\text{init}} \sim P(c_{\text{init}}|x)}[P(y_{\text{target}}|c_{\text{init}}, x)]\big) \\
&\leq \mathbb{E}_{c_{\text{init}} \sim P(c_{\text{init}}|x)}[-\log(P(y_{\text{target}}|c_{\text{init}}, x))]
\end{align*}

Minimizing the expected cross-entropy loss indirectly minimizes an upper bound on \(-\log(P(y_{\text{target}}|x))\), bringing us closer to maximizing \(P(y_{\text{target}}|x)\).


To approximate the expected value, we use Monte Carlo sampling with \(n\) samples of \(c_{\text{init}}\):

\begin{align*}
L_{\text{final}} &= \mathbb{E}_{c_{\text{init}} \sim P(c_{\text{init}}|x)}[-\log(P(y_{\text{target}}|c_{\text{init}}, x))] \\
&\approx \frac{1}{n} \sum_{i=1}^{n} -\log(P(y_{\text{target}}|c_{\text{init}, i}, x))
\end{align*}



Thus, minimizing the expected cross-entropy loss over sampled contexts is an effective approach to optimize text generation tasks.

\section{Warmup Generations Approach}
A general overview of our method is illustrated in Figure \ref{fig:workflow}. Unlike traditional SFT methods, where the loss is computed solely based on the input, our approach introduces an intermediate generation step. After receiving the input, the model first generates $n$ warmup sequences. The final loss is then computed as the average of $n$ individual losses, each conditioned on both the input and one of the generated warmup sequences. The pseudo-code outlining the implementation of our method is provided in 
\autoref{algo:1}.

\subsection{Implementation for Encoder-Decoder Models}
Encoder-decoder structured models have a clear separation between input and output. For this type of model, the input is processed through the encoder, and then $n$ $c_{init}$ are generated using beam search with sampling. Each generated $c_{init}$ is followed by a separator and fed into the decoder. Subsequently, the cross-entropy loss of $y_{target}$ is calculated $n$ times, conditioned on the input and each of the $n$ generated $c_{init}$. Finally, the average of these $n$ losses is taken as the final loss.

The inference process follows the same logic shown in Figure \autoref{fig:workflow}, given an input sequence, the model first generates a warmup sequence. This sequence is then concatenated with a separator and fed back into the beginning of the decoder. The model then generates the target sequence conditioning on both the original input and the generated warmup sequence. The final output consists of the tokens generated after the concatenated separator.

\subsection{Implementation for Decoder-Only Models}
Similar to encoder-decoder models, the input is first fed into the model, and $n$ $c_{init}$ are sampled using beam search. The input is then concatenated with $n$ $c_{init}$ sequences, followed by a separator, and fed back into the model. The final loss is the average cross-entropy loss of $y_{target}$ conditioned on the $n$ $c_{init}$ sequences and the input.

During inference, similar to encoder-decoder models, the warmup sequence is first generated and then appended to the input sequence, followed by a separator. The combined sequence is then fed back into the model to generate the target sequence. The final output consists of the tokens produced after the concatenated separator.
\subsection{Rationale Behind Using a Separator Between $c_{init}$ and $y_{target}$}
The inclusion of separators helps the model to distinguish the boundary between $c_{init}$ and $y_{target}$, preventing $y_{target}$ from being treated as a continuation of $c_{init}$. This enhances both the stability and efficiency of training. Since the separators are deterministically appended to the end of each $c_{init}$, the probability distributions of $c_{init}$ and $c_{init}$ followed by the separator remain the same. Additionally, since the model is rewarded based on the generation of $c_{init}$, the inclusion of separators does not disrupt this training process.

\begin{table}[t]
\centering
\resizebox{\columnwidth}{!}{%
\begin{tabular}{l c | c c}
\toprule
\textbf{Model} & \textbf{Warmup} &\textbf{Macro F1 } & \textbf{Accuracy} \\ 
 \midrule
\multirow{2}{*}{T5-base} & \checkmark & \textbf{50.00} & \textbf{50.06} \\ 
 & $\times$ &  49.16 & 49.19 \\ 
\multirow{2}{*}{T5-large}& \checkmark &  \textbf{55.50} & \textbf{55.62} \\ 
 & $\times$ & 54.46 & 54.54 \\ 

\multirow{2}{*}{Llama-3.2-1B}& \checkmark &  \textbf{32.63} & \textbf{33.55} \\ 
 & $\times$ & 30.12 & 31.70 \\ 
\midrule

\end{tabular}
}
\caption{Performance of each model on LogiQA2 for Macro F1 and Accuracy.}
\label{table:logiqa2}
\end{table}

\begin{table*}[htbp]
\centering
\scalebox{0.8}{
\begin{tabular}{l c c| r r r r| r r r r |c}
\toprule
\textbf{Model} & \textbf{Warmup} &\textbf{Metric} & \textbf{de-en} & \textbf{ru-en} & \textbf{zh-en} & \textbf{fr-en} & \textbf{en-de} & \textbf{en-ru} & \textbf{en-zh} & \textbf{en-fr} & \textbf{Avg} \\ 
\midrule
\multirow{6}{*}{mt5-base}       & \checkmark &\multirow{2}{*}{BLEU} &  \textbf{29.64}	&  \textbf{20.78}	&  \textbf{13.31}	&  \textbf{30.64}	&  \textbf{20.43}	&  \textbf{11.72}	&  \textbf{22.64}	&  \textbf{27.95}	&\textbf{22.14} \\ 
  & $\times$ & &    27.89	&   19.41	&  12.66	&   28.73	&   18.10&  10.64	&   20.95	&   26.14 & 20.57\\ 
      & \checkmark &\multirow{2}{*}{COMET} &  \textbf{83.12}	&  \textbf{78.19}	&  \textbf{77.05}	&  \textbf{82.99}	&  \textbf{75.11}	&  \textbf{69.64}&\textbf{75.89}	&  \textbf{76.89}& \textbf{77.36}\\ 
   & $\times$ &  &   82.21	&  77.60	&  75.50	&  81.95	&  72.88	& 68.13	&  74.45	&  75.60 &76.04\\ 
       & \checkmark &\multirow{2}{*}{ChrF++} &  \textbf{55.15}	&\textbf{46.48}	& \textbf{39.48}	&  \textbf{55.64}	&  \textbf{47.78}	&  \textbf{32.09}	&  \textbf{25.05}	&  \textbf{52.66}	 &\textbf{44.29}\\ 
       & $\times$ &  &    53.63& 45.57	&  38.01	&  53.84	&  45.43	&  30.71	&  23.44	&  50.87		&  42.69\\ 
\midrule
\multirow{6}{*}{mt5-large}       & \checkmark &\multirow{2}{*}{BLEU} &    \textbf{34.71}	&  \textbf{25.76}	&  \textbf{18.66}	& \textbf{35.63} &  \textbf{26.84}	&  \textbf{16.30}	&  \textbf{27.89}	&  \textbf{36.29}	&		  \textbf{27.76}\\ 
  & $\times$ & &    34.12	&  24.93	& 17.76&  34.84	& 24.91	&  14.29	&  27.29	&  30.39&26.07\\ 
      & \checkmark &\multirow{2}{*}{COMET} &  \textbf{86.74}	& \textbf{82.51}	&  \textbf{82.54}	& \textbf{86.37}		&   \textbf{82.67}	&   \textbf{77.43}	&  \textbf{82.19}	&  \textbf{83.50}		 &\textbf{82.99}\\ 
   & $\times$ & &    86.28&  82.30	&  81.83	&  86.12	&  81.47	& 74.43	&  82.18	& 80.91	 &81.94\\  
       & \checkmark &\multirow{2}{*}{ChrF++} &    \textbf{59.65}	&  \textbf{51.15}	& \textbf{45.45}& \textbf{59.39}		&  \textbf{53.19}	& \textbf{38.34}	&  \textbf{29.72}	&  \textbf{59.28}			&\textbf{49.52} \\ 
       & $\times$ &  &   58.75	& 50.80	& 44.21	&  58.87	&  51.44	&  34.47	&  29.17	&  54.33			& 47.75\\ 

\bottomrule
\end{tabular}
}
\caption{Performance on translation tasks with the comparison of the BLEU score, COMET score and ChrF++ score across different language pairs. For example, ``de-en'' denotes the source language to be German, and the target language to be English. }
\label{table:bleu_scores}
\end{table*}

\section{Experiments}

In this section, we present the tasks and corresponding datasets used,  the models selected for the experiments and the results obtained.

\subsection{Tasks and Datasets}
We evaluated our approach on three datasets spanning three tasks: FLORES~\cite{nllbteam2022languageleftbehindscaling} for testing, WMT for training for the translation task; LogiQA2~\cite{10174688} for logical reasoning multi-choice QA; and XSum~\cite{Narayan2018DontGM} for summarization;  
Specifically, we used WMT19 datasets \cite{barrault-etal-2019-findings} for the fine-tuning of de-en (en-de), ru-en (en-ru), and zh-en (en-zh) and the fr-en (en-fr) data from the WMT14 dataset \cite{bojar-etal-2014-findings}.

\subsection{Models}
We used T5-base (223M) and T5-large (738M) for summarization, T5-base, T5-large, and Llama-3.2-1B (1.24B) for multiple-choice logical reasoning, and mT5-base (582M) and mT5-large (1.23B) for translation. These models, covering both encoder-decoder and decoder-only architectures, serve as well-established benchmarks in their respective categories and are widely recognized for their effectiveness.

\subsection{Metrics}
We used the BLEU score \cite{papineni-etal-2002-bleu}, COMET score \footnote{The implementation of COMET was from huggingface.}  \cite{rei-etal-2020-comet}, and ChrF++ score \cite{popovic-2015-chrf} for translation. For logical reasoning multiple-choice, we used macro F1 and accuracy, and for summarization, we employed ROUGE score (1, 2, L) \cite{lin-2004-rouge} and BERTScore \footnote{The implementation of BERTScore was from huggingface, with the model type set to ``roberta-large''.} \cite{Zhang*2020BERTScore:}. 
\subsection{Experimental Settings}
For fine-tuning all tasks, we used a learning rate of 2e-5, with the warmup sequence's maximum sampled length capped at 8 tokens. Models were trained for 10 epochs. Due to computational constraints, we randomly selected 50,000 samples from the training set for fine-tuning in translation and summarization tasks. During fine-tuning, warmup sequences were generated using a beam size of 4, with 4 warmup sequences sampled per training sample for each loss calculation. 

For each task, we selected the checkpoint that achieved the highest metric score on the validation set and reported its performance on the test set. Specifically, for translation, checkpoints were selected based on the COMET score; for summarization, based on BERTScore; and for logical reasoning multiple-choice, based on the macro F1 score. For LogiQA2, each model was fine-tuned 3 times with different random seeds, and we report the average performance of the selected checkpoints.




\begin{table}[t]
\centering
\resizebox{\columnwidth}{!}{%
\begin{tabular}{llrrrr}
\toprule
 & &\multicolumn{3}{c}{\textbf{Rouge}} &\textbf{BERT}\  \\ 
\textbf{Model} & \textbf{Warmup} & \textbf{R1} & \textbf{R2} & \textbf{RL} &\textbf{Score}\  \\ 
\midrule
\multirow{2}{*}{T5-base}& \checkmark & \textbf{37.63} &15.51 &30.32& \textbf{90.50}\\ 
& $\times$ & 37.18 & 15.19 & 29.96 & 90.40\\ 
\midrule
\multirow{2}{*}{T5-large}& \checkmark & \textbf{40.67} &18.23& 33.21& \textbf{91.11}\\ 
 &  $\times$ & 40.21 &17.84 & 32.75& 91.05\\ 
\bottomrule
\end{tabular}
}
\caption{Performance on summarization tasks for Rouge (R1, R2 and RL) and BERTScore. }
\label{table:summarization}
\end{table}
\begin{table*}[ht!]
\centering
\renewcommand{\arraystretch}{1.6}
\footnotesize
\scalebox{0.9}{
\begin{tabular}{ccl}
\hline
\textbf{Lang Pair} & \textbf{Warmup Sequence} & \multicolumn{1}{p{0.5\textwidth}}{\textbf{\hspace{50mm}Content}} \\ \hline
\multicolumn{3}{c}{\textbf{Direct Core Phrases}} \\ \hline
\multirow{2}{*}{de-en} & \multirow{2}{*}{\parbox{0.2\textwidth}{\textcolor{softred}{``pyramid is the only one of''}}} & T: The Cheops \textcolor{softred}{pyramid is the only one of} the seven world wonders ... \\
                       &                                                                 & G: The Great \textcolor{softred}{Pyramid} at Giza \textcolor{softred}{is the only one of} the seven wonders ... \\ \hline
\multirow{2}{*}{fr-en} & \multirow{2}{*}{\parbox{0.2\textwidth}{\textcolor{softred}{``a British traveller in''}}}    & T: Similarly, \textcolor{softred}{a British traveller in} Spain could confuse ... \\
                       &                                                                 & G: Similarly, \textcolor{softred}{a British traveller in} Spain may mistake ... \\ \hline
\multirow{2}{*}{zh-en} & \multirow{2}{*}{\parbox{0.2\textwidth}{\textcolor{softred}{``a search on the Internet for''}}} & T: \textcolor{softred}{Search on the Internet for} a response to hostile environment courses ... \\
                       &                                                                 & G: \textcolor{softred}{A search of the Internet for} `Hostile environment course' ... \\ \hline
\multirow{2}{*}{ru-en} & \multirow{2}{*}{\parbox{0.2\textwidth}{\textcolor{softred}{``because there was no national''}}} & T: ... and \textcolor{softred}{because there was no national} executive or judicial power, ... \\
                       &                                                                 &G: ... and, \textcolor{softred}{because there was no national} executive or judiciary, ... \\ \hline
\multicolumn{3}{c}{\textbf{Similar Phrases}} \\ \hline
\multirow{2}{*}{de-en} & \multirow{2}{*}{\parbox{0.2\textwidth}{\textcolor{softblue}{``female travelers are recommended''}}}           & T: \textcolor{softblue}{Women: It is recommended that all women travelling} claim to be married, ... \\
                       &                                                                 & G: \textcolor{softblue}{Women: It is recommended that any women travellers} say that they are ... \\ \hline
\multirow{2}{*}{fr-en} & \multirow{2}{*}{\parbox{0.2\textwidth}{\textcolor{softblue}{``Please contact us directly''}}}                 & T: In all cases, you must \textcolor{softblue}{reserve by telephone directly} from the aircompany. \\
                       &                                                                 & G: In all cases, you must \textcolor{softblue}{book by phone directly} with the airline. \\ \hline
\multirow{2}{*}{zh-en} & \multirow{2}{*}{\parbox{0.2\textwidth}{\textcolor{softblue}{``shows a changing temperature''}}}               & T: The ultraviolet image \textcolor{softblue}{shows that the changes in the night temperature}  ... \\
                       &                                                                 & G: Infrared images \textcolor{softblue}{show that the temperature variations} from  night and day ... \\ \hline
\multirow{2}{*}{ru-en} & \multirow{2}{*}{\parbox{0.2\textwidth}{\textcolor{softblue}{``According to the Japanese nuclear''}}}          & T: \textcolor{softblue}{According to the nuclear authority of Japan}, radioactive cezai and ... \\
                       &                                                                 & G: \textcolor{softblue}{According to Japan's nuclear agency}, radioactive caesium and iodine ... \\ \hline
\end{tabular}
}
\caption{Examples of Direct Core Phrases and Similar Phrases for different language pairs. ``T'' indicates translations that are model-generate, and ``G'' indicates the golden label.}
\label{table:casestudy}
\end{table*}

\begin{table}[t]
\centering
\scalebox{0.8}{
\begin{tabular}{c| c c}
\hline
\textbf{Lang Pair} & \textbf{Devtest (\%)} & \textbf{Dev (\%)} \\ 
\hline
de-en                  & 55.05               & 56.57              \\ 
en-de                  & 32.71                  & 33.41              \\ 
fr-en                  & 40.94                 & 42.75              \\ 
en-fr                  & 34.42               & 32.97              \\ 
\midrule
zh-en                  & 43.91             & 42.90              \\ 
en-zh                  & 63.03                & 63.54             \\ 
ru-en                  & 46.13                  & 48.80              \\ 
en-ru                  & 33.51                 & 33.48              \\ 
\hline
\end{tabular}
}
\caption{Overlap rates of the warmup sequence for different language pairs on Flores200 devtest and dev datasets, generated by the mT5-base model.}

\label{tab:overlap-rate}
\end{table}

\subsection{Results and Discussions}
We put our experiment results in Table \ref{table:logiqa2}, \ref{table:bleu_scores}, and \ref{table:summarization}.
\paragraph{Warmup generations consistently enhance performance across tasks}
Across all three tasks, models utilizing warmup generations outperform those employing traditional SFT methods. The most significant gains are observed in translation tasks, where mT5-base achieves an average improvement of 1.57 BLEU, 1.32 COMET, and 1.60 ChrF++ scores across 8 language pairs.
For multiple-choice logical reasoning, the T5-base model trained with warmup generations achieves 0.84 higher macro F1 and 0.87 higher accuracy compared to models using traditional SFT. A similar trend is observed in summarization, where the T5-base model with warmup generations yields gains of 0.45 ROUGE-1, 0.32 R2, 0.36 RL, and 0.1 BERTScore. 
\paragraph{Performance gains are robust to increases in model size}
When scaling the models from base to large, we observe similar or even greater performance gains. In translation, mT5-large exhibits a greater average improvement than mT5-base, with a higher BLEU gain of 1.69, a COMET increase of 1.05 (slightly lower but still comparable), and a greater ChrF++ gain of 1.77.
For logical reasoning, T5-large benefits more from warmup generations than T5-base, achieving performance gains of 1.04 in Macro F1 and 1.08 in Accuracy. Similarly, in summarization, T5-large outperforms its counterpart without warmup generations, with improvements of 0.46, 0.39, and 0.46 in ROUGE-1, ROUGE-2, and ROUGE-L scores, respectively.  
\paragraph{Performance gain extends to decoder-only architecture}
The decoder-only model, Llama-3.2-1B, gains from warmup generations in multiple-choice logical reasoning, achieving performance improvements of 2.51 in Macro F1 and 1.85 in Accuracy.
\paragraph{Warmup generations improve lexical alignment more than semantic richness in summarization}
For summarization, using warmup generations achieves BERTScore increases by 0.06–0.1 points on a scale of 100, indicating that while warmup sequences enhance word selection and fluency, they do not significantly impact semantic richness. This suggests that, for summarization, warmup sequences help the model better mimic the word choices made in the reference summaries, leading to higher word-level alignment (ROUGE scores). However, they do not push the model to generate additional semantic content beyond what it would traditionally learn to extract through standard training, which explains the smaller improvement in BERTScore.

Overall, warmup sequences consistently improve performance across tasks, models, and architectures, but their effectiveness is task-dependent. 
For encoder-decoder models, 
logical reasoning benefits more in 
larger
models, 
while translation shows variable gains depending on the language pair. Summarization, in contrast, benefits uniformly across scales. 
While the decoder-only model, LLaMA-3.2-1B can leverage warmup sequences effectively, its relative performance remains lower than encoder-decoder models like T5.

\subsection{Ablation Studies}


\begin{figure*}
    \centering
    \includegraphics[width=0.48\textwidth, height=0.18\textheight]{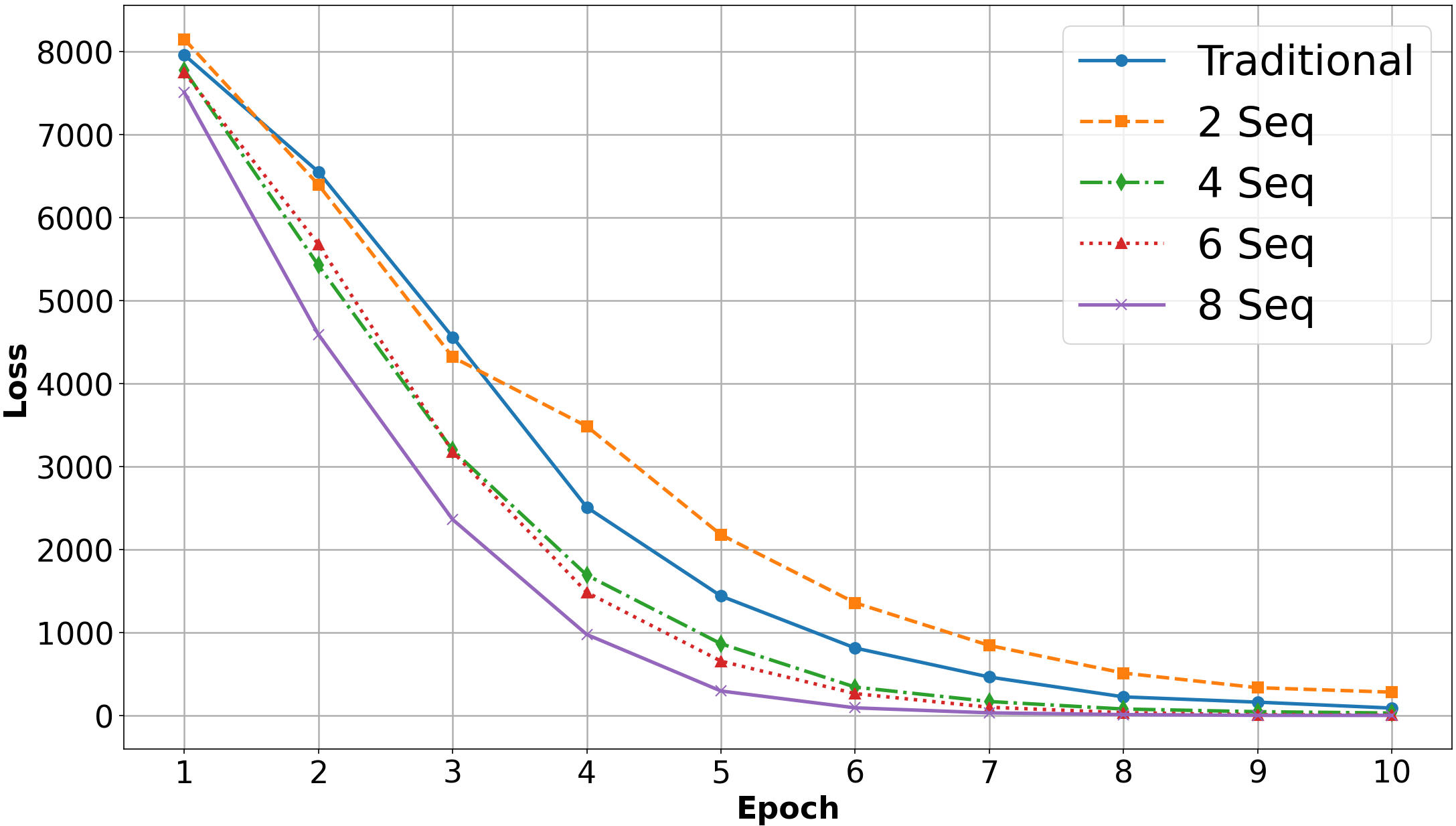}
    \includegraphics[width=0.48\textwidth, height=0.18\textheight]{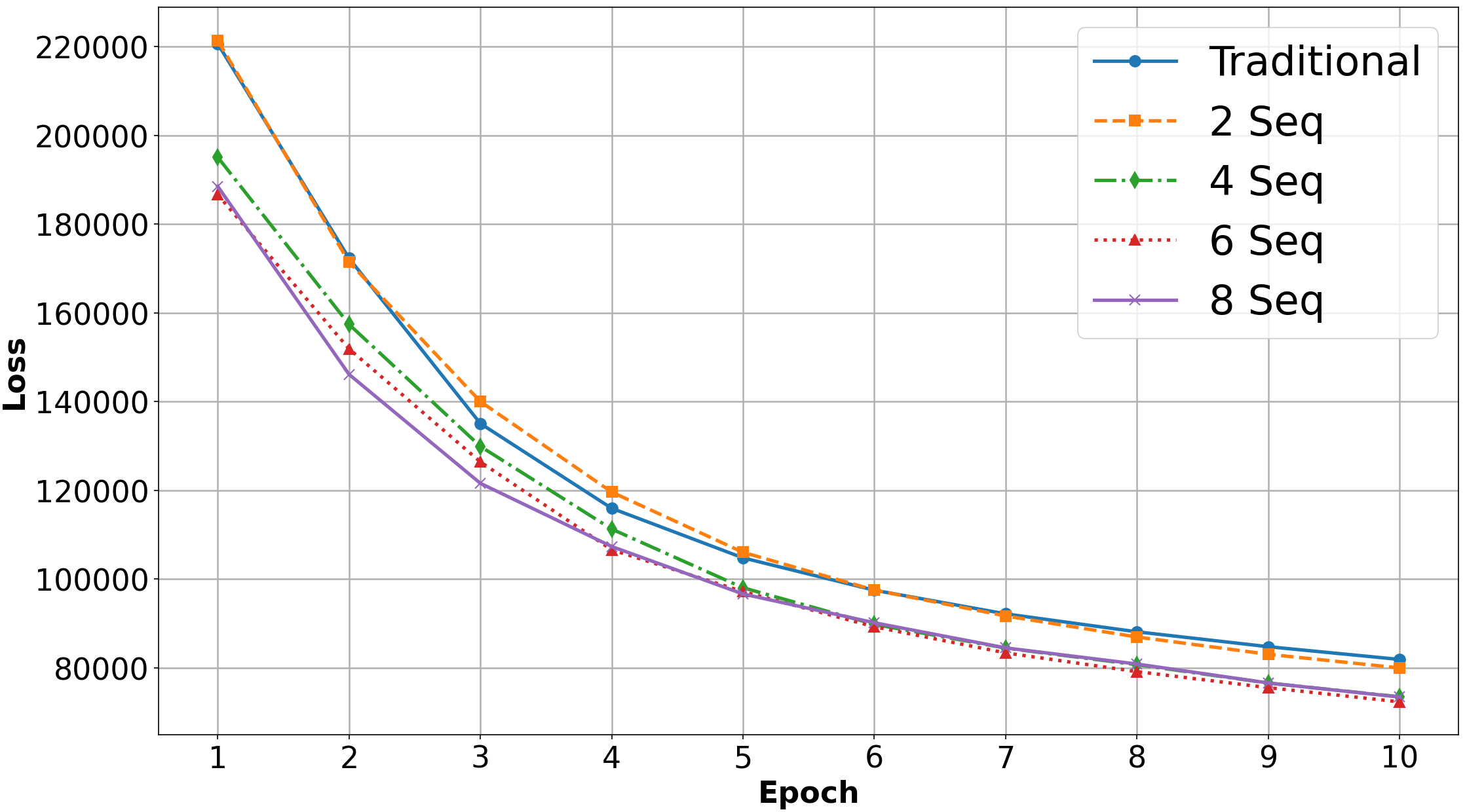}
    \caption{\textbf{Models' training loss after each epoch of fine-tuning}. The left figure represents T5-base on the LogiQA2 dataset, while the right graph represents mT5-base on the WMT19 dataset for the de-en language pair.}
    \label{fig:loss_logiqa2}
\end{figure*}

\subsubsection{Number of Samples}

\begin{table}[t]
\centering
\resizebox{\columnwidth}{!}{%
\begin{tabular}{c|ccc|cc}
\toprule
 & \multicolumn{3}{c}{\textbf{de-en translation}} & \multicolumn{2}{c}{\textbf{LogiQA2}}\  \\ 
\textbf{Sample} & \textbf{BLEU} & \textbf{COMET} & \textbf{ChrF++} & \textbf{Macro F1} &\textbf{Accuracy}\  \\ 
\midrule
4 & 29.64  & 83.12 & 55.15 & 50.00 & 50.06 \\ 
2 & 28.51  & 82.45 & 54.07 & 50.09 & 50.17 \\ 
6 & \textbf{29.70} & \textbf{83.24} & 55.26 & 50.19 & 50.19  \\ 
8 & 29.62  & 83.06  & \textbf{55.38} & \textbf{50.99} & \textbf{50.95} \\ 
\bottomrule
\end{tabular}
}
\caption{Performance of mT5-base (de-en translation) and T5-base (LogiQA2) models trained with varying numbers of warmup sequences sampled during training. }
\label{table:seqnum_study}
\end{table}

To assess the impact of the number of sampled warmup sequences during training, we analyze both training loss trends and test set performance across different sample numbers.

As shown in Figure \ref{fig:loss_logiqa2}, increasing the number of sampled warmup sequences generally accelerates convergence and reduces final training loss. 
However, the differences between 4 and 6 sequences for LogiQA2, and 4, 6, and 8 sequences for de-en translation are relatively small, suggesting that beyond a certain threshold, additional samples do not significantly reduce training loss further.

The results for LogiQA2 in Table \ref{table:seqnum_study} indicate that increasing the number of samples does not lead to strictly monotonic improvements. The default setting of 4 samples achieves 50.00 Macro F1 and 50.06 Accuracy while increasing to 6 samples provides only a slight improvement (50.19 Macro F1 and 50.19 Accuracy). However, moving from 6 to 8 samples results in the largest jump, with Macro F1 increasing to 50.99 and Accuracy to 50.95. Notably, reducing the number of samples to 2 still achieves 50.09 Macro F1 and 50.17 accuracy,
this suggests that even a small number of warmup samples can provide meaningful improvements.

The results for de-en translation exhibit a similar trend. While increasing the number of samples improves translation quality up to 6 sequences, beyond this point, the gains become marginal or even slightly decrease. Specifically, BLEU improves from 28.51 (2 samples) to 29.70 (6 samples) but slightly drops to 29.62 with 8 samples. COMET score follows a similar pattern, increasing from 82.45 to 83.24 before slightly decreasing to 83.06, while ChrF++ continues improving slightly, though the changes are minimal beyond 6 samples.

These findings suggest that while increasing the number of warmup sequences reduces training loss and improves test-time performance, there exists an optimal range—such as 6 to 8 samples for de-en translation—beyond which the benefits diminish. The initial improvements stem from better approximations of the probability distribution of warmup sequences, leading to more effective learning. However, adding too many samples can introduce redundancy or increased variance, limiting further performance gains.

\subsection{Qualitative Analysis}
We performed a qualitative analysis of the translation task to investigate the role of warmup sequences. 
As the results shown in Table \ref{table:casestudy}, we found that these warmup sequences can be primarily categorized into two types:
\begin{itemize}
\item Direct Core Phrases: These warmup sequences can be directly identified in both the labels and the generated outputs.
\item Similar Phrases: Expressions that are semantically similar to important components in the labels and outputs.
\end{itemize}


For example, ``a British traveller in''  and ``a series of events that'' can be directly found in both the output and ground-truth labels. This indicates that for certain scenarios, initial states function as core-information extractors, guiding the model to generate outputs focusing on these core concepts.

Meanwhile, in other cases, the warmup sequences are more semantically related rather than exact phrase matches. For instance, ``when they are in danger'' is semantically related to ``they perceive a threat'' in the label, and ``Please contact us directly'' aligns with ``book by phone directly with the airline.'' In such scenarios, the initial states serve as semantic guides, enabling the model to generate outputs that capture the intended meaning without relying on exact phrase matching.

To further evaluate this dual role, we calculated the overlapping rate of the words in the initial states that also appear in the labels in Table \ref{tab:overlap-rate}.
As we can see, ``X-eng'' all achieve a overlapping rate over 40\%, with ``de-en'' reaching 56.57\% at the most.
On ``Eng-X'', the overlapping rate also reach at least 32.71\% on ``en-de'', and get to the highest on ``en-zh'' with the rate of 63.53\%. 
This means that generally, over 40\% of the words in the warmup sequence could be found in the ground-truth, indicating a strong alignment between the initial states and the ground-truth data. 
By acting as both direct extractors and semantic interpreters, the initial states ensure the generated outputs remain closely aligned with the intended semantics and structure of the target language.

\section{Conclusions}
In this work, we introduced a task-agnostic framework with theoretical proof and derivation, for improving sequence-to-sequence learning through warmup generations, where models learn to generate intermediate sequences to enhance final output quality. Unlike traditional approaches, our method learns intermediate steps in an unsupervised manner, improving performance across diverse tasks without requiring task-specific annotations.
Experiments demonstrate that warmup sequences consistently benefit both encoder-decoder and decoder-only models across different sizes. Analysis reveals that warmup sequences aid generation by extracting key phrases and providing semantically related guidance, resulting in more fluent and contextually accurate outputs. Additionally, increasing the number of sampled warmup sequences accelerates convergence and enhances test-time performance, though gains diminish beyond a certain threshold.
However, the performance gains vary across tasks and architectures, highlighting the need for further investigation into how different task types and model structures influence warmup effectiveness.
Overall, by introducing and demonstrating the effectiveness of warmup sequences across multiple seq2seq tasks, this work lays the groundwork for further research into leveraging intermediate generations to enhance model training and generation.

\section*{Limitations}
While our proposed framework demonstrates improvements across various tasks, there are several limitations to address.
The first is increased training time. The framework relies on sampling multiple initial states during training, introducing computational overhead compared to traditional supervised fine-tuning methods. This can make training more resource-intensive, particularly for large-scale datasets or deployment in constrained environments. Future work could explore more efficient sampling strategies or adaptive selection methods to mitigate this cost.
Another limitation is that warmup sequences primarily enhance the model's lexical-level understanding rather than deeper reasoning or structural-level improvements. 
As shown in our experiments, warmup generation aids the model in selecting key phrases and improving word choice, but it does not explicitly introduce or infer new knowledge beyond what is present in the input. Future research could explore how warmup sequences might be adapted to facilitate higher-level abstraction or knowledge augmentation, potentially bridging gaps in implicit reasoning.
Finally, our framework has not been tested on decoder-only models for generative tasks. While experiments on LogiQA2 demonstrate improvements for decoder-only architectures, the application of warmup sequences to open-ended text generation (e.g., summarization or translation) in decoder-only models remains unexplored. This poses potential challenges, as decoder-only models lack explicit input-output alignments found in sequence-to-sequence tasks, making it unclear whether warmup sequences would be equally effective. Investigating warmup generation within causal language models is an important direction for future work.

\bibliography{anthology,custom}

\appendix
\label{sec:templates}
\section{Selection of Seperator}
The separator token for T5 and mT5 was set to `` || '', as this symbol is rarely used in natural text, making it an ideal choice for separating different parts of the sequence. 
\section{Warmup Sequence for Summarization and Logical Reasoning}

The warmup sequences for summarization follow a similar pattern to those in translation, predominantly consisting of either Direct Core Phrases or Similar Phrases. For logical reasoning, the warmup sequence is identical to the target sequence, representing only the final letter choice that indicates the predicted answer.



\end{document}